\documentclass[sigconf,9pt]{acmart}
\AtBeginDocument{%
  }
    \usepackage{enumitem}

\copyrightyear{2026}
\acmYear{2026}
\setcopyright{cc}
\setcctype{by}
\acmConference[ACM Sustainability Week Companion '26]{ACM Sustainability Week 2026}{June 22--25, 2026}{Banff, AB, Canada}
\acmBooktitle{ACM Sustainability Week 2026 (ACM Sustainability Week Companion '26), June 22--25, 2026, Banff, AB, Canada}
\acmDOI{10.1145/3765611.3815392}
\acmISBN{979-8-4007-2199-1/2026/06}

\usepackage{acmart-taps}  
\usepackage{booktabs}  
\usepackage{multirow} 
\usepackage{makecell}
\usepackage{placeins} 
\usepackage{graphicx}
\usepackage{subcaption}
\usepackage{dsfont}

\begin{document}

\title{Investigating Calibration Challenges in Probabilistic Electricity Price Forecasting}


\author{Jan Niklas Lettner}
\authornote{Jan Niklas Lettner and Hadeer El Ashhab contributed equally to this work.}
\email{jan.lettner@kit.edu}
\orcid{0009-0004-5995-4607}
\affiliation{%
\department{Institute for Automation and Applied
Informatics}
  \institution{Karlsruhe Institute of Technology}
  \city{Karlsruhe}
  \state{Baden-Württemberg}
  \country{Germany}
}

\author{Hadeer El Ashhab}\authornotemark[1]
\email{hadeer.elashhab@kit.edu}
\orcid{0009-0000-5132-0504}
\affiliation{%
\department{Institute for Automation and Applied
Informatics}
  \institution{Karlsruhe Institute of Technology}
  \city{Karlsruhe}
  \state{Baden-Württemberg}
  \country{Germany}
}

\author{Benjamin Schäfer}
\email{benjamin.schaefer@kit.edu}
\orcid{0000-0003-1607-9748}
\affiliation{%
\department{Institute for Automation and Applied
Informatics}
  \institution{Karlsruhe Institute of Technology}
  \city{Karlsruhe}
  \state{Baden-Württemberg}
  \country{Germany}
}

\renewcommand{\shortauthors}{Lettner et al.}

\begin{abstract}
As renewable energy integration increases market volatility, probabilistic electricity price forecasting has become essential for effective risk management. However, current--proper--scoring rules often prioritize forecast sharpness at the expense of calibration, leading to overconfident and statistically unreliable uncertainty estimates. This work highlights the critical gap between theoretical scoring and practical calibration, demonstrating that models can become mere proxies for deterministic forecasts when reliability is neglected. We conclude that future research must shift toward calibration-aware objectives and architectures to ensure the distributional integrity of energy market forecasts.
\end{abstract}

\ccsdesc[500]{Applied computing~Physical sciences and engineering}
\ccsdesc[500]{Computing methodologies~Machine learning}
\keywords{Probabilistic Forecasting, Calibration, Quantile Regression, Time Series, Time Series Forecasting, Energy, Electricity Prices, Electricity Price Forecasting, Cross-border}


\maketitle

\section{Introduction}
\label{sec:introduction}

\begin{figure}[t ]
    \centering
    \includegraphics[width=0.8\linewidth]{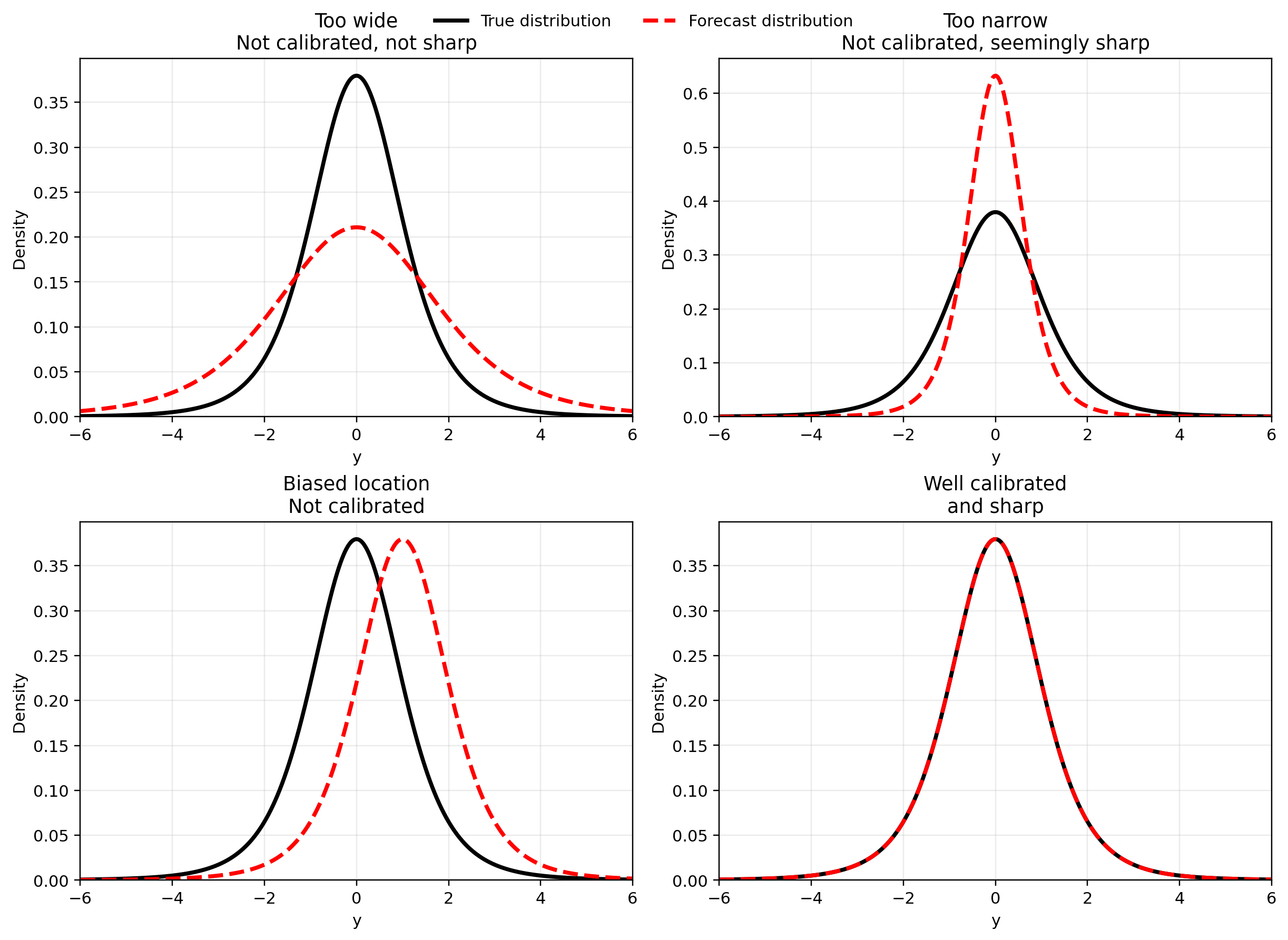}
    \caption{Illustrative forecast examples showing underdispersion, overdispersion, bias, and ideal calibration. Note that calibration is a collective property requiring statistical diagnostics (e.g., PIT histograms)} 
    \label{fig:calibration_variations}
\end{figure}

In recent years, forecasting research has increasingly shifted from point forecasts toward probabilistic forecasts \cite{maciejowskaForecastingElectricityPrices2022} as it outperforms point estimates by quantifying predictive uncertainty, providing better decision support under volatility. This shift necessitates the use of proper scoring rules that account for both sharpness and calibration--the fundamental paradigm being to maximize sharpness subject to calibration \cite{gneitingProbabilisticForecasting2014}. Calibration refers to the statistical consistency between the predicted probabilities and the actual observed frequencies; essentially, if an event is predicted with a $90\%$ probability, it should occur $90\%$ of the time in the long run. Sharpness, on the other hand, refers to the concentration of the predictive distributions; a "sharper" forecast provides narrower prediction intervals, offering more precise information for the decision-maker. If a model focuses exclusively on sharpness while neglecting calibration, it risks becoming a mere proxy for deterministic forecasting--providing precise-looking estimates that lack statistical reliability and fail to represent the true underlying risks. Figure~\ref{fig:calibration_variations} provides a schematic illustration of these concepts by contrasting ideal, underdispersed, overdispersed, and biased forecast distributions.

Quantile Regression Averaging (QRA) \cite{nowotarskiComputingElectricitySpot2015} generates probabilistic forecasts by treating point forecasts and explanatory variables as inputs for separate quantile regressions. These models are optimized via pinball loss, a proper scoring rule minimized only when the forecast matches the true underlying distribution. Despite the theoretical appeal of proper scoring rules, empirical evidence shows they can still result in miscalibration. Notably, Si et al. observed this in normalizing flows trained with negative log-likelihood (NLL) \cite{siSemiAutoregressiveEnergyFlows2023}, while Chung et al. found that pinball loss can prioritize sharpness over calibration \cite{chungPinballLossQuantile2021}. This leads to overconfident forecasts, violating the principle of maximizing sharpness subject to calibration. This work aims to highlight the critical gap between existing proper scoring rules and the requirement for metrics that more effectively penalize miscalibration. By addressing this gap, we hope that forecasts remain statistically sound and well-calibrated rather than merely superficially precise. Building on climate change mitigation initiatives that advocate for the integration of highly volatile renewable energy sources \cite{putzFeasibilityForecastingHighly2024}, probabilistic electricity price forecasting (PEPF) has become essential for both market participants and grid operators alike. To this end, the data selection, feature engineering, and model architectures employed in this work are all inspired by prior work in PEPF \cite{lettner2026assessingperformanceefficiencytradeofffoundation}.

\section{Methods}
\label{sec:methods}
Our analysis uses the same hourly EU electricity prices as in earlier work \cite{lettner2026assessingperformanceefficiencytradeofffoundation}. To predict day-ahead prices, the model uses the previous 168 hours as input and outputs forecasts for the next 24 hours. Our pipeline uses a two-stage design: first, NHITS generates stochastic point forecasts via Monte Carlo dropout. These forecasts, along with calendar features, are then passed to a QRA head to produce final probabilistic predictions, following the architecture of Lettner et al. \cite{lettner2026assessingperformanceefficiencytradeofffoundation}. We compare two objectives for quantile estimation: the standard pinball loss and the calibration-oriented loss proposed by Chung et al. \cite{chungPinballLossQuantile2021}. The latter includes a calibration component that checks whether the empirical coverage of the predicted quantiles matches their nominal coverage, by shifting quantile predictions upward or downward when needed, as well as a sharpness term to avoid excessively wide prediction intervals. Under the standard pinball loss, the QRA head consists of a separate linear regression model for each horizon and quantile. When applying the loss of Chung et al., we instead aim to more closely follow their original architecture by using a neural network with two hidden layers of 64 units each and a learning rate of $10^{-3}$. In this setting, the model is trained separately for each horizon. The weighting parameter controlling the trade-off between sharpness and calibration is tuned in the main paper \cite{chungPinballLossQuantile2021}. In our experiments, we set it to 0.5, thereby assigning equal importance to both objectives. We investigate group calibration to improve predictive reliability. While average calibration only requires quantile accuracy across the entire dataset, group calibration demands this accuracy within specific subgroups. Following Chung et al., we enforce this by alternating between standard and group-based training batches. We set the alternating frequency to 2, meaning the model switches batching schemes every other step. We also train a Normalizing Flow model using masked autoregressive layers, conditioned on latent representations from a Transformer encoder-decoder. While such models usually minimize NLL, Si et al. suggest that NLL alone may be insufficient and report better performance using alternative loss functions \cite{siSemiAutoregressiveEnergyFlows2023}.

\section{Results and Discussion}
\label{sec:preliminary_results}

\aptLtoX{\begin{table*}[t]
    \centering
\begin{tabular}{ll}
        \begin{tabular}{lcc}
            \hline
            Used Loss & CRPS & ECE \\
            \hline
            Pinball & 16.87 & 0.107 \\
            Chung  & 21.74 & 0.273 \\
            Chung + G. & 21.74 & 0.274 \\
            \hline
            \multicolumn{3}{l}{\scriptsize G.\ denotes group batching by weekday/hour.}\\
            \multicolumn{3}{l}{\textbf{(a) Results for NHITS+QRA}}
        \end{tabular} &
       \begin{tabular}{lcc}
            \hline
            Used Loss & CRPS & ECE \\
            \hline
            NLL & 31.51 & 0.219 \\
            CRPS & 24.87 & 0.145 \\
            ES & 25.08 & 0.156 \\
            \hline
            \multicolumn{3}{l}{\scriptsize ES denotes Energy Score.}\\
\multicolumn{3}{l}{\textbf{(b) Results for Normalizing flows}}
        \end{tabular}
        \end{tabular}
    \caption{Comparison of probabilistic forecasting models in terms of CRPS and ECE on the test dataset.}
    \label{tab:comparison_models}
\end{table*}}{\begin{table}[t]
    \centering

    \begin{subtable}[t]{0.48\linewidth}
        \centering
        \begin{tabular}{lcc}
            \hline
            Used Loss & CRPS & ECE \\
            \hline
            Pinball & 16.87 & 0.107 \\
            Chung  & 21.74 & 0.273 \\
            Chung + G. & 21.74 & 0.274 \\
            \hline
            \multicolumn{3}{l}{\scriptsize G.\ denotes group batching by weekday/hour.}
        \end{tabular}
        \caption{Results for NHITS+QRA}
        \label{tab:nhitsqra}
    \end{subtable}
    \hfill
    \begin{subtable}[t]{0.48\linewidth}
        \centering
        \begin{tabular}{lcc}
            \hline
            Used Loss & CRPS & ECE \\
            \hline
            NLL & 31.51 & 0.219 \\
            CRPS & 24.87 & 0.145 \\
            ES & 25.08 & 0.156 \\
            \hline
            \multicolumn{3}{l}{\scriptsize ES denotes Energy Score.}
        \end{tabular}
        \caption{Results for Normalizing flows}
        \label{tab:nf}
    \end{subtable}

    \caption{Comparison of probabilistic forecasting models in terms of CRPS and ECE on the test dataset.}
    \label{tab:comparison_models}
\end{table}}


As shown in Table~\ref{tab:comparison_models}, the pinball loss  of NHITS+QRA remains a strong baseline. The loss by Chung et al. performed substantially worse in continuous ranked probability score (CRPS) and expected calibration error (ECE), with group batching yielding no improvement. This discrepancy may stem from the non-i.i.d. nature of electricity prices; unlike standard regression datasets, temporal dependencies and seasonal shifts may undermine the effectiveness of Chung et al.'s calibration objective. Alternatively, our use of separate models for each horizon and quantile under the pinball setup may offer greater flexibility than a single, more expressive model. For the normalizing flow model, training with CRPS yielded the best CRPS and ECE results. These findings demonstrate that, while various losses are theoretically motivated, the choice of scoring rule significantly impacts empirical calibration in practice. Probability integral transform (PIT) histograms further reveal the nature of these errors, with results for both models  showing clear signs of bias or overconfidence.


\section{Conclusion and Outlook}
\label{sec:conclusions_outlook}

The integration of renewable energy sources necessitates PEPF to manage increasing market volatility. However, generating probabilistic forecasts that are both sharp and well-calibrated remains a challenge. While sharpness can be captured through standard deterministic metrics, the true value of a probabilistic model lies in its calibration--the extent to which predictive distributions reflect actual outcomes. We show that current "proper" scoring rules often fail to ensure this calibration, frequently favoring sharpness at the expense of reliability. Ultimately, because the primary goal of probabilistic forecasting  is to provide a dependable measure of uncertainty, future research should shift focus from maximizing sharpness to developing calibration-aware objectives and architectures that prioritize the distributional integrity of the forecasts.

\bibliographystyle{ACM-Reference-Format}
\bibliography{ref1}

\end{document}